\documentclass{article}

\usepackage{amsmath}
\usepackage[utf8]{inputenc} 
\usepackage[T1]{fontenc}    
\usepackage{lmodern}
\usepackage{url}            
\usepackage{booktabs}       
\usepackage{amsfonts}       
\usepackage{nicefrac}       
\usepackage{microtype}      
\usepackage{graphicx} 
\usepackage{xcolor}
\usepackage{subfigure}
\usepackage{mathrsfs}
\usepackage{bm}
\usepackage[margin=1.0in]{geometry}

\newcommand{\citee}[1]{\cite{#1}}

\usepackage{amsthm}
\usepackage{amsmath}
\usepackage{amssymb}
\usepackage{color}
\usepackage{pgfplots, pgfplotstable}

\newcommand{\comment}[1]{}

\renewcommand{\vec}[1]{\boldsymbol{\mathbf{#1}}}
\def\btheta{\vec{\theta}}

\def\calY{\mathcal{Y}}

\def\modelp{{P}_{\theta}} 

\def\objml{\mathcal{O}_{\mathrm{ML}}}

\def\objrl{\mathcal{O}_{\mathrm{RL}}}
\def\objmixed{\mathcal{O}_{\mathrm{Mixed}}}

\def\dataset{\mathcal{D}}

\def\int{\mathrm{int}}

\newcommand{\reward}[2]{r(#1, #2)}

\def\LSTM{\mathrm{LSTM}}

\title{Google's Neural Machine Translation System: Bridging the Gap between Human and Machine Translation}

\author{Yonghui Wu, Mike Schuster, Zhifeng Chen, Quoc V. Le, Mohammad Norouzi\\
        \texttt{yonghui,schuster,zhifengc,qvl,mnorouzi@google.com} \and
        Wolfgang Macherey,
        Maxim Krikun,
        Yuan Cao,
        Qin Gao,
        Klaus Macherey,\\
        Jeff Klingner,
        Apurva Shah,
        Melvin Johnson,
        Xiaobing Liu,
        \L{}ukasz Kaiser,\\
        Stephan Gouws,
        Yoshikiyo Kato,
        Taku Kudo,
        Hideto Kazawa,
        Keith Stevens,\\
        George Kurian,
        Nishant Patil,
        Wei Wang,
        Cliff Young,
        Jason Smith,
        Jason Riesa,\\
        Alex Rudnick,
        Oriol Vinyals,
        Greg Corrado,
        Macduff Hughes,
        Jeffrey Dean\\
}

\date{}
\begin{document}
\maketitle

\begin{abstract}
Neural Machine Translation (NMT) is an end-to-end learning approach for
automated translation, with the potential to overcome many of the weaknesses
of conventional phrase-based translation systems.
Unfortunately, NMT systems are known to be computationally expensive both in
training and in translation inference --
sometimes prohibitively so in the case of very large data sets and large models.
Several authors have also charged that NMT systems lack robustness,
particularly when input sentences contain rare words.
These issues have hindered NMT's use in practical deployments and services,
where both accuracy and speed are essential.
In this work, we present GNMT, Google's Neural Machine Translation system,
which attempts to address many of these issues.
Our model consists of a deep LSTM network with 8 encoder and 8 decoder layers using residual connections as well as
attention connections from the decoder network to the encoder.
To improve parallelism and therefore decrease training time,
our attention mechanism connects the bottom layer of the decoder to the top
layer of the encoder.
To accelerate the final translation speed, we employ low-precision arithmetic
during inference computations.
To improve hand\-ling of rare words, we divide words into a limited set of
common sub-word units (``wordpieces'') for both input and output.
This method provides a good balance between the flexibility of 
``character''-delimited models and the efficiency of ``word''-delimited models,
naturally handles translation of rare words, and ultimately improves the
overall accuracy of the system.
Our beam search technique employs a length-normalization procedure and
uses a coverage penalty, which encourages generation of an output
sentence that is most likely to cover all the words in the source sentence.
To directly optimize the translation BLEU scores, we consider refining the
models by using reinforcement learning, but we found that the improvement in
the BLEU scores did not reflect in the human evaluation.
On the WMT'14 English-to-French and English-to-German benchmarks,
GNMT achieves competitive results to state-of-the-art.
Using a human side-by-side evaluation on a set of isolated simple sentences,
it reduces translation errors by an average of 60\% compared to
Google's phrase-based production system.

\end{abstract}
\section{Introduction}
\label{intro}
Neural Machine Translation
(NMT)~\cite{sutskever2014sequence,BahdanauCB15} has recently been
introduced as a promising approach with the potential of addressing
many shortcomings of traditional machine translation systems.
The strength of NMT lies in its ability to learn directly, in an
end-to-end fashion, the mapping from input text to associated output text.
Its architecture typically consists of two recurrent neural networks (RNNs), one
to consume the input text sequence and one to generate translated output text.
NMT is often accompanied by an attention mechanism~\cite{BahdanauCB15}
which helps it cope effectively with long input sequences.

An advantage of Neural Machine Translation is that it sidesteps many
brittle design choices in traditional phrase-based machine
translation~\cite{koehn2003statistical}. In practice, however, NMT systems 
used to be worse in accuracy than phrase-based translation systems,
especially when training on very large-scale datasets as used for the very 
best publicly available translation systems.
Three inherent weaknesses of Neural Machine Translation are responsible for this
gap: its slower training and inference speed, ineffectiveness in dealing with
rare words, and sometimes  
failure to translate all words in the source sentence. Firstly, it generally
takes a considerable amount of time and computational resources to 
train an NMT system on a large-scale translation dataset, thus slowing the rate 
of experimental turnaround time and innovation. For inference they are generally
much slower than phrase-based systems due to the large number of parameters
used.
Secondly, NMT lacks robustness in translating rare words. Though this
can be addressed in principle by training a ``copy model'' to mimic a
traditional alignment model~\cite{luong2015addressing}, or by using the
attention mechanism to copy rare words~\cite{jean2015using}, these approaches are
both unreliable at scale, since the quality of the alignments varies across
languages, and the latent alignments produced by the attention
mechanism are unstable when the network is deep. Also, simple copying
may not always be the best strategy to cope with rare words, for example when
a transliteration is more appropriate. Finally,
NMT systems sometimes produce output sentences
that do not translate all parts of the input sentence -- in other
words, they fail to completely ``cover'' the input, which can result in
surprising translations.

This work presents the design and implementation of GNMT, a production NMT
system at Google, that aims to
provide solutions to the above problems. In our implementation, the
recurrent networks are Long Short-Term Memory (LSTM)
RNNs~\cite{hochreiter1997long,gers2000learning}. Our LSTM RNNs have 8
layers, with residual connections between layers to encourage gradient
flow~\cite{he2015deep}. For parallelism, we connect the attention from
the bottom layer of the decoder network to the top layer of the
encoder network. To improve inference time, we employ low-precision
arithmetic for inference, which is further accelerated by special
hardware (Google's Tensor Processing Unit, or TPU).  To effectively
deal with rare words, we use sub-word units (also known as
``wordpieces'') \cite{wordpiece_schuster} for inputs and outputs in
our system. Using wordpieces gives a good balance between the
flexibility of single characters and the efficiency of full words for
decoding, and also sidesteps the need for special treatment of unknown
words.  Our beam search technique includes a length normalization procedure to
deal efficiently with the problem of comparing hypotheses of different
lengths during decoding, and a coverage penalty to encourage the model
to translate all of the provided input.

Our implementation is robust, and performs well on a range of datasets
across many pairs of languages without the need for language-specific
adjustments. Using the same implementation, we are able to achieve
results comparable to or better than previous state-of-the-art
systems on standard benchmarks, while delivering great improvements over
Google's phrase-based production translation system.
Specifically, on WMT'14 English-to-French, our single model
scores 38.95 BLEU, an improvement of 7.5 BLEU from a single model
without an external alignment model reported in~\cite{luong2015addressing} and an improvement of 1.2 BLEU from a single model
without an external alignment model reported in~\cite{DBLP:journals/corr/ZhouCWLX16}. 
Our single model is also comparable to a single model in~\cite{DBLP:journals/corr/ZhouCWLX16}, 
while not making use of any alignment model as being used in~\cite{DBLP:journals/corr/ZhouCWLX16}.
Likewise on WMT'14 English-to-German,
our single model scores 24.17 BLEU, which is 3.4 BLEU better
than a previous competitive baseline~\cite{buck2014n}. On production data, our
implementation is even more effective.  Human evaluations show that GNMT has
reduced translation errors by 60\% compared to our previous phrase-based system
on many pairs of languages: English $\leftrightarrow$ French, English
$\leftrightarrow$ Spanish, and English $\leftrightarrow$ Chinese.
Additional experiments suggest the quality of the resulting translation system
gets closer to that of average human translators.

\section{Related Work}
\label{relwork}
Statistical Machine Translation (SMT) has been the dominant translation
paradigm for
decades~\cite{Brown:1988:SAL:991635.991651,brown1990statistical,Brown:1993:MSM:972470.972474}.
Practical implementations of SMT are generally phrase-based systems (PBMT)
which translate sequences of words or phrases where the lengths may
differ~\cite{koehn2003statistical}.


Even prior to the advent of direct Neural Machine Translation,
neural networks have been used as a component within SMT systems with some success.
Perhaps one of the most notable attempts involved the use of a joint language
model to learn phrase representations~\cite{devlin2014fast} which yielded an
impressive improvement when combined with phrase-based translation.
This approach, however, still makes use of phrase-based translation systems
at its core, and therefore inherits their shortcomings.
Other proposed approaches for learning phrase representations~\cite{ChoMGBSB14}
or learning end-to-end translation with neural 
networks~\cite{kalchbrenner2013recurrent} offered encouraging hints, but 
ultimately delivered worse overall accuracy compared to standard 
phrase-based systems.

The concept of end-to-end learning for machine translation has been
attempted in the past (e.g.,~\cite{chrisman1991learning}) with limited
success. Following seminal papers in the
area~\cite{sutskever2014sequence,BahdanauCB15}, NMT translation quality has
crept closer to the level of phrase-based translation systems for
common research benchmarks. Perhaps the first successful attempt at surpassing
phrase-based translation was described in~\cite{luong2015addressing}.
On WMT'14 English-to-French, this system achieved a 0.5 BLEU improvement
compared to a state-of-the-art phrase-based system.

Since then, many novel techniques have been proposed to further
improve NMT: using an attention mechanism to deal with rare
words~\cite{jean2015using}, a mechanism to model translation
coverage~\cite{TuLLLL16}, multi-task and semi-supervised training to
incorporate more data~\cite{dong2015multi,luong2015multi}, a character
decoder~\cite{chung2016character}, a character
encoder~\cite{DBLP:journals/corr/Costa-JussaF16}, subword
units~\cite{SennrichHB15} also to deal with rare word outputs,
different kinds of attention
mechanisms~\cite{luong-pham-manning:2015:EMNLP}, and sentence-level
loss minimization~\cite{ShenCHHWSL15,RanzatoCAZ15}.
While the translation accuracy of these systems has been encouraging, systematic
comparison with large scale, production quality phrase-based translation systems
has been lacking.

\section{Model Architecture}
\label{model architecture}
\begin{figure}[h]
\begin{center}
\centerline{\includegraphics[width=\textwidth]{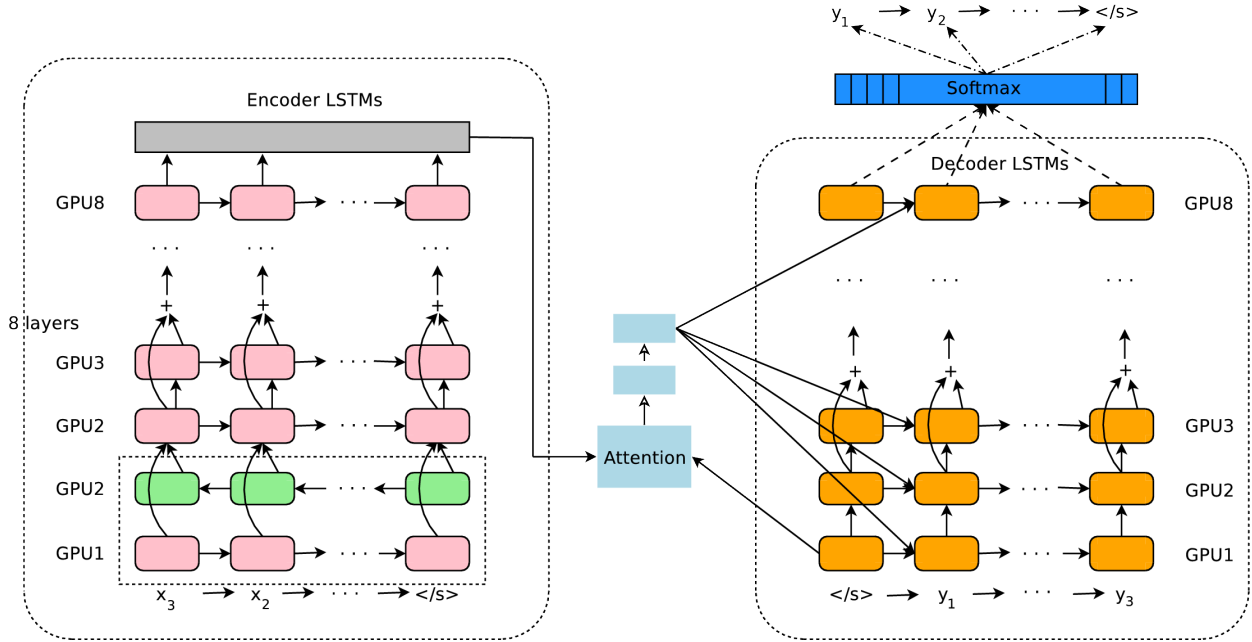}}
\caption{The model architecture of GNMT, Google's Neural Machine
  Translation system. On the left is the encoder network, on the right is the
  decoder network, in the middle is the attention module. The bottom
  encoder layer is bi-directional: the pink nodes gather information
  from left to right while the green nodes gather information from
  right to left. The other layers of the encoder are uni-directional. Residual
  connections start from the layer third from the bottom in the encoder and
  decoder. The model is partitioned into multiple GPUs to speed
  up training. In our setup, we have 8 encoder LSTM layers (1
  bi-directional layer and 7 uni-directional layers), and 8 decoder
  layers. With this setting, one model replica is partitioned 8-ways
  and is placed on 8 different GPUs typically belonging to one host
  machine. During training, the bottom bi-directional encoder layers
  compute in parallel first. Once both finish, the
  uni-directional encoder layers can start computing, each on a separate GPU. 
  To retain as much parallelism as possible during running the decoder layers, 
  we use the 
  bottom decoder layer output only for obtaining recurrent attention context,
  which is sent directly to all the remaining
  decoder layers. The softmax layer is also partitioned and placed on
  multiple GPUs. Depending on the output vocabulary size we either have
  them run on the same GPUs as the encoder and decoder networks, or
  have them run on a separate set of dedicated GPUs.}
\label{main_figure}
\end{center}
\end{figure}

Our model (see Figure~\ref{main_figure}) follows the common
sequence-to-sequence learning framework~\cite{sutskever2014sequence} with
attention~\cite{BahdanauCB15}. It has three components:
an encoder network, a decoder network, and an attention network. The
encoder transforms a source sentence into a list of vectors, one vector per input symbol.  Given
this list of vectors, the decoder produces one symbol at a time, until
the special end-of-sentence symbol (EOS) is produced. The encoder and decoder
are connected through an attention module which allows the decoder to
focus on different regions of the source sentence during the course of
decoding.



For notation, we use bold lower case to denote vectors (e.g., $\mathbf{v,
o_i}$), bold upper case to represent matrices (e.g., $\mathbf{U, W}$), cursive
upper case to represent sets (e.g., $\mathscr{V, T}$), capital letters to represent sequences (e.g. $X$, $Y$), and lower
case to represent individual symbols in a sequence, (e.g., $x_1$, $x_2$).

Let $(X, Y)$ be a source and target sentence pair.  Let
$X = x_1, x_2, x_3, ..., x_{M}$ be the
sequence of $M$ symbols in the source sentence and let 
$Y = y_1, y_2, y_3, ..., y_{N}$ be the sequence of
$N$ symbols in the target sentence. The encoder is simply a function of
the following form:

\begin{equation}
\label{eq-encoder}
\mathbf{x_1, x_2, ..., x_{M}} = EncoderRNN(x_1, x_2, x_3, ..., x_{M})
\end{equation}
In this equation, $\mathbf{x_1, x_2, ..., x_{M}}$ is a
list of fixed size vectors. The number of members in the list is the
same as the number of symbols in the source sentence ($M$ in this
example). Using the chain rule the conditional probability of the sequence
$P(Y | X)$ can be decomposed as:

\begin{equation}
\label{p_y_given_x}
	\begin{split}
	P(Y | X) &= P(Y | \mathbf{x_1, x_2, x_3, ..., x_{M}}) \\
	&= \prod_{i=1}^{N}P(y_i | y_0, y_1, y_2, ..., y_{i-1}; \mathbf{x_1, x_2, x_3, ..., x_{M}})
	\end{split}
\end{equation}
where $y_0$ is a special ``beginning of sentence'' symbol that is
prepended to every target sentence.

During inference we calculate the probability of the next symbol given
the source sentence encoding and the decoded target sequence so far:
\begin{equation}
\label{eq-decoder}
	P(y_i|y_0, y_1, y_2, y_3, ...,
        y_{i-1}; \mathbf{x_1}, \mathbf{x_2}, \mathbf{x_3},...,
        \mathbf{x_{M}})
\end{equation}
Our decoder is implemented as a combination of an RNN network and a
softmax layer. The decoder RNN network produces a hidden state
$\mathbf{y_i}$ for the next symbol to be predicted, which then goes
through the softmax layer to generate a probability distribution over candidate output symbols.

In our experiments we found that for NMT systems to achieve good accuracy, 
both the encoder and decoder RNNs have to be deep enough to capture subtle
irregularities in the source and target languages. This observation is
similar to previous observations that deep LSTMs significantly outperform 
shallow LSTMs~\cite{sutskever2014sequence}. In that work, each
additional layer reduced perplexity by nearly 10\%. Similar to
\cite{luong2015addressing}, we use a deep stacked Long Short Term
Memory (LSTM)~\cite{hochreiter1997long} network for both the encoder
RNN and the decoder RNN.

Our attention module is similar to~\cite{BahdanauCB15}. More
specifically, let $\mathbf{y}_{i-1}$ be the decoder-RNN output from
the past decoding time step (in our implementation, we use the output from
the bottom decoder layer).  Attention context $\mathbf{a}_i$
for the current time step is computed according to the following formulas:
\begin{equation}
\label{atten}
	\begin{split}
	s_t &= AttentionFunction(\mathbf{y}_{i-1}, \mathbf{x}_{t}) \quad  \forall t,   \quad  1 \leq t \leq M \\
	p_t &= \exp(s_t) / \sum_{t=1}^{M}\exp(s_t) \quad  \quad \forall t, \quad 1 \leq t \leq M \\
	\mathbf{a}_i &= \sum_{t=1}^{M} p_t . \mathbf{x}_t
	\end{split}
\end{equation}
where $AttentionFunction$ in our implementation is a feed forward network with 
one hidden layer.

\subsection{Residual Connections}
\label{Residual Connections}
As mentioned above, deep stacked LSTMs often give better accuracy over 
shallower models.  
However, simply stacking more layers of LSTM works only to a certain number of
layers, beyond which the network becomes too slow and difficult to
train, likely due to exploding and vanishing gradient problems
\cite{DBLP:journals/corr/abs-1211-5063,Hochreiter01gradientflow}. In
our experience with large-scale translation tasks, simple stacked LSTM layers
work well up to 4
layers, barely with 6 layers, and very poorly beyond 8 layers.

\begin{figure}[h!]
\begin{center}
\centerline{
\includegraphics[width=\textwidth]{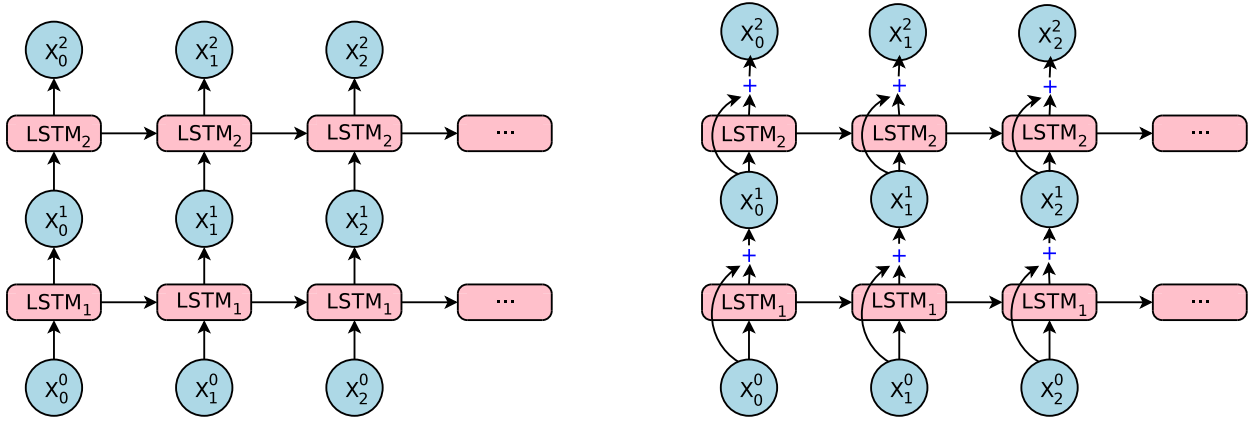}
}
\caption{The difference between normal stacked LSTM and our stacked
  LSTM with residual connections. On the left: simple stacked LSTM
  layers~\citee{sutskever2014sequence}. On the right: our
  implementation of stacked LSTM layers with residual
  connections. With residual connections, input to the bottom LSTM
  layer ($\mathbf{x_i^0}$'s to $\LSTM_1$) is element-wise added to the
  output from the bottom layer ($\mathbf{x_i^1}$'s). This sum is then
  fed to the top LSTM layer ($\LSTM_2$) as the new input.}
\label{residual_connection}
\end{center}
\end{figure}


Motivated by the idea of modeling differences between an intermediate layer's
output and the targets, which has shown to work well for many projects in the
past \cite{Fahlman90thecascade-correlation,he2015deep,DBLP:journals/corr/SrivastavaGS15},
we introduce residual connections
among the LSTM layers in a stack (see Figure~\ref{residual_connection}). 
More concretely, let $\LSTM_i$ and $\LSTM_{i+1}$ be the $i$-th and
$(i+1)$-th LSTM layers in a stack, whose parameters are
$\mathbf{W}^i$ and $\mathbf{W}^{i+1}$ respectively. At the
$t$-th time step, for the stacked LSTM without residual
connections, we have:

\begin{equation}
	\label{LSTM_stack}
	\begin{split}
	\mathbf{c}_t^i,\mathbf{m}_t^i &= \LSTM_i(\mathbf{c}_{t-1}^i, \mathbf{m}_{t-1}^i, \mathbf{x}_t^{i-1}; \mathbf{W}^i) \\
	\mathbf{x}_t^{i} & =  \mathbf{m}_t^i \\
	\mathbf{c}_t^{i+1},\mathbf{m}_t^{i+1} &= \LSTM_{i+1}(\mathbf{c}_{t-1}^{i+1}, \mathbf{m}_{t-1}^{i+1}, \mathbf{x}_t^{i}; \mathbf{W}^{i+1})
	\end{split}
\end{equation}
where $\mathbf{x}_t^i$ is the input to $\LSTM_i$ at time step $t$,
and $\mathbf{m}_t^i$ and $\mathbf{c}_t^i$ are the hidden states and memory
states of $\LSTM_i$ at time step $t$, respectively. 

With residual connections between $\LSTM_{i}$ and $\LSTM_{i+1}$, the
above equations become:

\begin{equation}
	\label{LSTM_stack_residual}
	\begin{split}
	\mathbf{c}_t^i,\mathbf{m}_t^i &= \LSTM_i(\mathbf{c}_{t-1}^i, \mathbf{m}_{t-1}^i,	\mathbf{x}_t^{i-1}; \mathbf{W}^i) \\
	\mathbf{x}_t^{i} & =  \mathbf{m}_t^i + \mathbf{x}_t^{i-1} \\
	\mathbf{c}_t^{i+1},\mathbf{m}_t^{i+1} &= \LSTM_{i+1}(\mathbf{c}_{t-1}^{i+1}, \mathbf{m}_{t-1}^{i+1}, \mathbf{x}_t^{i}; \mathbf{W}^{i+1})
	\end{split}
\end{equation}
Residual connections greatly improve the gradient flow in the backward
pass, which allows us to train very deep encoder and decoder
networks. In most of our experiments, we use 8 LSTM layers for the encoder
and decoder, though residual connections can allow us to train
substantially deeper networks (similar to what was observed
in~\cite{DBLP:journals/corr/ZhouCWLX16}).




%

\subsection{Bi-directional Encoder for First Layer}

For translation systems, the information required to translate certain words
on the output side can appear anywhere on the source side. 
Often the source side information is approximately left-to-right, similar to 
the target side, but depending on the language pair the information for
a particular output word can be distributed and even be split up in certain
regions of the input side.

\begin{figure}[h!]
\begin{center}
\centerline{
\includegraphics[width=0.6\textwidth]{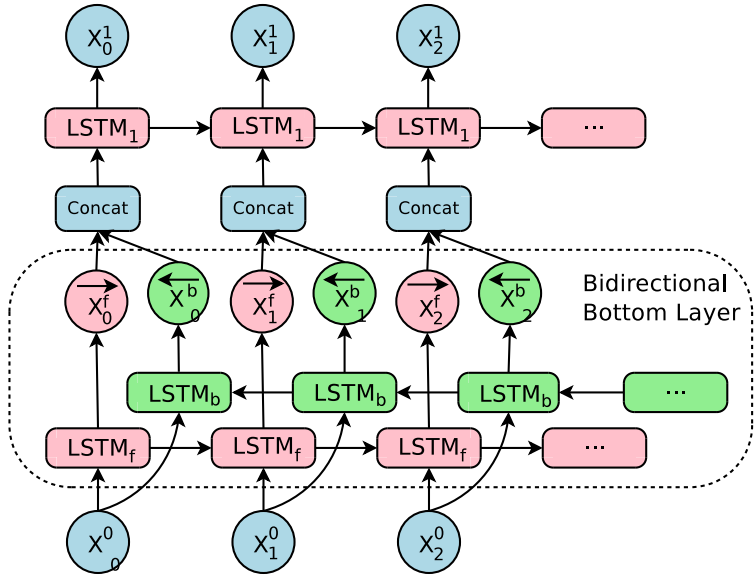}
}
\caption{The structure of bi-directional connections in the first layer
  of the encoder. LSTM layer $\LSTM_f$ processes information from left
  to right, while LSTM layer $\LSTM_b$ processes information from right
  to left. Output from $\LSTM_f$ and $\LSTM_b$ are first concatenated
  and then fed to the next LSTM layer $\LSTM_1$.}
\label{bidirectional_rnn}
\end{center}
\end{figure}

To have the best possible context at each point in the encoder network
it makes sense to use a bi-directional
RNN~\cite{Schuster:1997:BRN:2198065.2205129} for the encoder, which
was also used in~\cite{BahdanauCB15}.  To allow for maximum possible 
parallelization during computation (to be
discussed in more detail in section \ref{Model Parallelism}),
bi-directional connections are only used for the bottom encoder layer -- all
other encoder layers are uni-directional. Figure
\ref{bidirectional_rnn} illustrates our use of bi-directional LSTMs at
the bottom encoder layer. The layer $\LSTM_f$ processes the source
sentence from left to right, while the layer $\LSTM_b$ processes the
source sentence from right to left. Outputs from $\LSTM_f$
($\overrightarrow{\mathbf{x_t^f}}$) and $\LSTM_b$
($\overleftarrow{\mathbf{x_t^b}}$) are first concatenated and then fed
to the next layer $\LSTM_1$.


\subsection{Model Parallelism}
\label{Model Parallelism}

Due to the complexity of our model, we make use of both model
parallelism and data parallelism to speed up training. Data
parallelism is straightforward: we train $n$ model replicas
concurrently using a Downpour SGD algorithm \cite{downpoursgd}. The
$n$ replicas all share one copy of model parameters, with each replica
asynchronously updating the parameters using a combination of Adam
\cite{DBLP:journals/corr/KingmaB14} and SGD algorithms.  In our experiments, 
$n$ is often around 10. Each replica works on a mini-batch of $m$
sentence pairs at a time, which is often 128 in our experiments.

In addition to data parallelism, model parallelism is used to improve
the speed of the gradient computation on each replica. The encoder and
decoder networks are partitioned along the depth dimension and are
placed on multiple GPUs, effectively running each layer on a different GPU.
Since all but the first encoder layer are uni-directional, layer $i+1$ 
can start its computation before layer $i$ is fully finished, which improves
training speed.
The softmax layer is also partitioned, with
each partition responsible for a subset of symbols in the output
vocabulary.  Figure \ref{main_figure} shows more details of how
partitioning is done.

Model parallelism places certain constraints on the model
architectures we can use. For example, we cannot afford to have
bi-directional LSTM layers for all the encoder layers, since doing so
would reduce parallelism among subsequent layers, as each layer would
have to wait until both forward and backward directions of the previous
layer have finished. This would effectively constrain us to make use of
only 2 GPUs in parallel (one for the forward direction and one for the
backward direction). For the attention portion of the model, we chose to align the
bottom decoder output to the top encoder output to maximize
parallelism when running the decoder network. Had we aligned the top decoder
layer to the top encoder layer, we would have removed all parallelism
in the decoder network and would not benefit from using more than one
GPU for decoding.


\section{Segmentation Approaches}
Neural Machine Translation models often operate with fixed word
vocabularies even though translation is fundamentally an open vocabulary problem
(names, numbers, dates etc.).  There are two broad categories of
approaches to address the translation of out-of-vocabulary (OOV)
words. One approach is to simply \textbf{copy} rare words from source to target
(as most rare words are names or numbers where the correct translation
is just a copy), either based on the attention
model~\cite{jean2015using}, using an external alignment
model~\cite{luong2015addressing}, or even using a more complicated
special purpose pointing
network~\cite{DBLP:journals/corr/GulcehreANZB16}. Another broad
category of approaches is to use \textbf{sub-word units}, e.g.,
chararacters~\cite{DBLP:journals/corr/ChungCB16}, mixed
word/characters~\cite{DBLP:journals/corr/LuongM16}, or more intelligent
sub-words~\cite{SennrichHB15}.

\subsection{Wordpiece Model}
\label{wordpiece Model}
Our most successful approach falls into the second category (sub-word units), and we
adopt the wordpiece model (WPM) implementation initially developed to
solve a Japanese/Korean segmentation problem for the Google speech
recognition system~\cite{wordpiece_schuster}. This approach is completely
data-driven and guaranteed to generate a deterministic segmentation
for any possible sequence of characters.  It is similar to
the method used in \cite{SennrichHB15} to deal with rare words in
Neural Machine Translation.

For processing arbitrary words, we first break words into wordpieces
given a trained wordpiece model. Special word boundary symbols are
added before training of the model such that the original word
sequence can be recovered from the wordpiece sequence without
ambiguity. At decoding time, the model first produces a wordpiece
sequence, which is then converted into the corresponding word
sequence.

Here is an example of a word sequence and the corresponding wordpiece sequence:
\begin{itemize}
\item {\bf Word}: Jet makers feud over seat width with big orders at stake
\item {\bf wordpieces}: \_J et \_makers \_fe ud \_over \_seat \_width \_with \_big \_orders \_at \_stake
\end{itemize}

In the above example, the word ``Jet'' is broken into two wordpieces
``\_J'' and ``et'', and the word ``feud'' is broken into two
wordpieces ``\_fe'' and ``ud''. The other words remain as single
wordpieces. ``\_'' is a special character added to mark the
beginning of a word.

The wordpiece model is generated using a data-driven approach to
maximize the language-model likelihood of the training data, given an
evolving word definition. Given a training corpus and a number of
desired tokens $D$, the optimization problem is to select $D$
wordpieces such that the resulting corpus is minimal in the number of
wordpieces when segmented according to the chosen wordpiece model. Our
greedy algorithm to this optimization problem is similar
to~\cite{SennrichHB15} and is described in more detail in
\cite{wordpiece_schuster}. Compared to the original implementation used in
\cite{wordpiece_schuster}, we use a special symbol only at the
beginning of the words and not at both ends.  We also cut the number
of basic characters to a manageable number depending on the data
(roughly 500 for Western languages, more for Asian languages) and map
the rest to a special unknown character to avoid polluting the given
wordpiece vocabulary with very rare characters. We find that using a total vocabulary of between 8k and 32k wordpieces achieves
both good accuracy (BLEU scores) and fast decoding
speed across all pairs of language pairs we have tried.

As mentioned above, in translation it often makes sense to copy rare
entity names or numbers directly from the source to the target. To
facilitate this type of direct copying, we always use a shared
wordpiece model for both the source language and target
language. Using this approach, it is guaranteed that the same string
in source and target sentence will be segmented in exactly the same
way, making it easier for the system to learn to copy these tokens.

Wordpieces achieve a balance between the flexibility of characters and
efficiency of words.
We also find that our models get better overall BLEU scores when using
wordpieces -- possibly due to the fact that our models now deal
efficiently with an essentially infinite vocabulary without resorting to
characters only. The latter would make the average lengths of the input and output 
sequences much longer, and therefore would require more computation. 


\subsection{Mixed Word/Character Model}
\label{Mixed Word/Character Model}
A second approach we use is the mixed word/character model.
As in a word model, we keep a fixed-size word vocabulary.
However, unlike in a conventional word model where OOV words are collapsed 
into a single UNK symbol, we convert OOV words into the sequence of its 
constituent characters.
Special prefixes are prepended to the characters, to 1) show the location of
the characters in a word, and 2) to distinguish them from normal in-vocabulary
characters. There are three
prefixes: {\tt <B>,<M>, and <E>}, indicating beginning of the word, middle 
of the word and end of the word, respectively. For example, let's assume the 
word {\tt Miki} is not in the vocabulary. It will be preprocessed into a
sequence of special tokens: {\tt <B>M <M>i <M>k <E>i}. The process is
done on both the source and the target sentences. During decoding, the
output may also contain sequences of special tokens. With the
prefixes, it is trivial to reverse the tokenization to the original words as 
part of a post-processing step.

\section{Training Criteria}
\label{sec:RL}
Given a dataset of parallel text containing $N$ input-output sequence
pairs, denoted $\dataset \equiv \left\{(X^{(i)}, Y^{*(i)})\right\}_{i=1}^N$,
standard maximum-likelihood training aims at maximizing the sum of log
probabilities of the ground-truth outputs given the corresponding
inputs,
\begin{equation}
	\objml(\btheta) = \sum_{i=1}^N \log \modelp(Y^{*(i)} \mid
	X^{(i)})~.
\label{eq:objml}
\end{equation}
The main problem with this objective is that it does not reflect the
task reward function as measured by the BLEU score in translation. Further,
this objective does not explicitly encourage a ranking among \textit{incorrect}
output sequences -- where outputs with higher BLEU scores should still obtain
higher probabilities under the model -- since incorrect outputs are never 
observed during training. In other words, using maximum-likelihood
training only, the model will not learn to be robust to errors made during 
decoding since they are never observed, which is quite a mismatch between 
the training and testing procedure.


Several recent papers~\cite{RanzatoCAZ15,ShenCHHWSL15,rml}
have considered different ways of incorporating the task reward into
optimization of neural sequence-to-sequence models.
In this work, we also attempt to refine a model pre-trained on the
maximum likelihood objective to directly optimize for the task reward.
We show that, even on large datasets, refinement of state-of-the-art
maximum-likelihood models using task reward improves the results
considerably.

We consider model refinement using the expected reward objective (also
used in~\cite{RanzatoCAZ15}), which can be expressed as
\begin{equation}
	\objrl(\btheta) = \sum_{i=1}^N \sum_{Y \in \calY}
	\modelp(Y \mid
	X^{(i)})~\reward{Y}{Y^{*(i)}}.
\label{eq:objrl}
\end{equation}
Here, $\reward{Y}{Y^{*(i)}}$ denotes the per-sentence
score, and we are computing an expectation over all of the output
sentences $Y$, up to a certain length.

The BLEU score has some undesirable properties when used for single
sentences, as it was designed to be a corpus measure. We therefore use a slightly
different score for our RL experiments which we call the ``GLEU score''.
For the GLEU score, we record all sub-sequences of 1, 2, 3 or 4 tokens
in output and target sequence (n-grams). We then compute a recall, which is the
ratio of the number of matching n-grams to the number of total n-grams in the
target (ground truth) sequence, and a precision, which is the ratio of the number of
matching n-grams to the number of total n-grams in the generated output sequence.
Then GLEU score is simply the minimum of recall and precision. This GLEU
score's range is always between 0 (no matches) and 1 (all match) and
it is symmetrical when switching output and target. According to our
experiments, GLEU score correlates quite well with the BLEU metric on a
corpus level but does not have its drawbacks for our per sentence reward objective.

As is common practice in reinforcement learning, we subtract the mean reward
from $\reward{Y}{Y^{*(i)}}$ in equation \ref{eq:objrl}. The
mean is estimated to be the sample mean of $m$ sequences
drawn independently from distribution $\modelp(Y \mid
X^{(i)})$. In our implementation, $m$ is set to be 15.
To further stabilize training, we optimize a linear combination of ML
(equation \ref{eq:objml}) and RL (equation \ref{eq:objrl}) objectives 
as follows:
\begin{equation}
	\objmixed(\btheta) = \alpha * \objml(\btheta) + \objrl(\btheta)
\label{eq:mixed}
\end{equation}
$\alpha$ in our implementation is typically set to be $0.017$.

In our setup, we first train a model using the maximum likelihood
objective (equation \ref{eq:objml}) until convergence. We then
refine this model using a mixed maximum likelihood and expected reward
objective (equation
\ref{eq:mixed}), until BLEU score on a development set is no longer improving. 
The second step is optional.


\section{Quantizable Model and Quantized Inference}
One of the main challenges in deploying our Neural Machine Translation
model to our interactive production translation service is that it is 
computationally intensive at inference, making low latency translation 
difficult, and high volume deployment computationally expensive. 
Quantized inference using reduced precision arithmetic is
one technique that can significantly reduce the cost of inference for these 
models, often providing efficiency improvements on the same computational devices. 
For example, in \cite{DBLP:journals/corr/WuLWHC15}, it is demonstrated
that a convolutional neural network model can be sped up by a factor of 4-6
with minimal loss on classification accuracy on the ILSVRC-12
benchmark. In \cite{DBLP:journals/corr/LiL16}, it is demonstrated that
neural network model weights can be quantized to only three states,
-1, 0, and +1.

Many of those previous studies \cite{DBLP:journals/corr/GuptaAGN15,DBLP:journals/corr/HanMD15,DBLP:journals/corr/WuLWHC15,DBLP:journals/corr/LiL16}
however mostly focus on CNN models with
relatively few layers. Deep LSTMs with long sequences pose
a novel challenge in that quantization errors can be significantly
amplified after many unrolled steps or after going through a deep
LSTM stack.


In this section, we present our approach to speed up inference with
quantized arithmetic. Our solution is tailored towards the hardware
options available at Google. To reduce quantization errors, additional
constraints are added to our model during training so that it is quantizable 
with minimal impact on the output of the model. That is, once a
model is trained with these additional constraints, it can be subsequently
quantized without loss to translation quality. Our experimental results suggest
that those additional constraints do not hurt model convergence nor the quality
of a model once it has converged.


Recall from equation \ref{LSTM_stack_residual} that in an LSTM stack
with residual connections there are two accumulators: $\mathbf{c}_t^i$
along the time axis and $\mathbf{x}_{t}^i$ along the depth axis. In
theory, both of the accumulators are unbounded, but in practice, we
noticed their values remain quite small. For quantized inference, we
explicitly constrain the values of these accumulators to be within
[-$\delta$, $\delta$] to guarantee a certain range that can be used for
quantization later. The forward computation of an LSTM stack with
residual connections is modified to the following:

\begin{equation}
	\label{LSTM_stack_residual_quatized}
	\begin{split}
	\mathbf{c'}_{t}^i,\mathbf{m}_{t}^i &= \LSTM_i(\mathbf{c}_{t-1}^i, \mathbf{m}_{t-1}^i, \mathbf{x}_t^{i-1}; \mathbf{W}^i) \\
        \mathbf{c}_{t}^i &= \max(-\delta, \min(\delta, \mathbf{c'}_{t}^i)) \\
        \mathbf{x'}_t^{i} & =  \mathbf{m}_t^i + \mathbf{x}_t^{i-1} \\
		\mathbf{x}_t^{i} & =  \max(-\delta, \min(\delta, \mathbf{x'}_t^{i})) \\
	    \mathbf{c'}_{t}^{i+1},\mathbf{m}_{t}^{i+1} &= \LSTM_{i+1}(\mathbf{c}_{t-1}^{i+1}, \mathbf{m}_{t-1}^{i+1}, \mathbf{x}_t^{i}; \mathbf{W}^{i+1}) \\
        \mathbf{c}_{t}^{i+1} &= \max(-\delta, \min(\delta, \mathbf{c'}_{t}^{i+1}))
	\end{split}
\end{equation}
Let us expand $\LSTM_i$ in equation \ref{LSTM_stack_residual_quatized}
to include the internal gating logic.  For brevity, we drop all the
superscripts $i$.

\begin{equation}
	\begin{split}
	\label{LSTM_internal}
		\mathbf{W} &= [\mathbf{W}_1, \mathbf{W}_2,  \mathbf{W}_3, \mathbf{W}_4,  \mathbf{W}_5,  \mathbf{W}_6,  \mathbf{W}_7, \mathbf{W}_8] \\
		\mathbf{i}_{t} &= \text{sigmoid}(\mathbf{W}_1 \mathbf{x}_t + \mathbf{W}_2 \mathbf{m}_t)  \\
		\mathbf{i'}_{t} &= \tanh(\mathbf{W}_3 \mathbf{x}_t + \mathbf{W}_4 \mathbf{m}_t)  \\
		\mathbf{f}_{t} &= \text{sigmoid}(\mathbf{W}_5 \mathbf{x}_t + \mathbf{W}_6 \mathbf{m}_t)  \\
		\mathbf{o}_{t} &= \text{sigmoid}(\mathbf{W}_7 \mathbf{x}_t + \mathbf{W}_8 \mathbf{m}_t) \\
		\mathbf{c}_{t} &= \mathbf{c}_{t-1} \odot \mathbf{f}_{t} + \mathbf{i'}_{t} \odot \mathbf{i}_{t} \\
		\mathbf{m}_{t} &= \mathbf{c}_{t} \odot \mathbf{o}_{t}
	\end{split}
\end{equation}
When doing quantized inference, we replace all the floating point
operations in equations \ref{LSTM_stack_residual_quatized} and
\ref{LSTM_internal} with fixed-point integer operations with either
8-bit or 16-bit resolution. The weight matrix $\mathbf{W}$ above is
represented using an 8-bit integer matrix $\mathbf{WQ}$ and a float
vector $\mathbf{s}$, as shown below:

\begin{equation}
	\label{weight_quantization}
	\begin{split}
		\mathbf{s}_i &= \max(\text{abs}(\mathbf{W}[i,:])) \\
		\mathbf{WQ}[i,j] &= \text{round}(\mathbf{W}[i, j] /
		\mathbf{s}_i \times 127.0)
	\end{split}
\end{equation}
All accumulator values ($\mathbf{c}_{t}^i$ and $\mathbf{x}_{t}^i$) are represented using
16-bit integers representing the range $[-\delta, \delta]$. All matrix
multiplications (e.g., $\mathbf{W}_1 \mathbf{x}_t$,
$\mathbf{W}_2 \mathbf{m}_t$, etc.) in equation \ref{LSTM_internal}
are done using 8-bit integer multiplication accumulated into larger
accumulators. All other operations, including all the activations
($\text{sigmoid}$, $\tanh$) and elementwise operations ($\odot$, $+$)
are done using 16-bit integer operations.

We now turn our attention to the log-linear softmax layer. During training,
given the decoder RNN network output $\mathbf{y_t}$, we compute the probability
vector $\mathbf{p_t}$ over all candidate output symbols as follows:

\begin{equation}
	\label{softmax_quantization}
	\begin{split}
		\mathbf{v_t} &= \mathbf{W_s} * \mathbf{y_t} \\
		\mathbf{v_t'} &= \max(-\gamma, \min(\gamma, \mathbf{v_t})) \\
		\mathbf{p_t} &= softmax(\mathbf{v_t'})
	\end{split}
\end{equation}
In equation \ref{softmax_quantization}, $\mathbf{W_s}$ is the weight
matrix for the linear layer, which has the same number of rows as the
number of symbols in the target vocabulary with each row corresponding
to one unique target symbol.  $\mathbf{v}$ represents the raw logits, which are
first clipped to be between $-\gamma$ and $\gamma$ and then normalized
into a probability vector $\mathbf{p}$. Input $\mathbf{y_t}$ is
guaranteed to be between $-\delta$ and $\delta$ due to the
quantization scheme we applied to the decoder RNN. The clipping range
$\gamma$ for the logits $\mathbf{v}$ is determined empirically, and in
our case, it is set to $25$. In quantized inference, the weight matrix
$\mathbf{W_s}$ is quantized into 8 bits as in
equation \ref{weight_quantization}, and the matrix multiplication is done using 
8 bit arithmetic. The calculations within the $softmax$ function and the 
attention model are not quantized during inference.

%

\begin{figure}[h!]
\begin{center}
\centerline{\includegraphics[width=\textwidth]{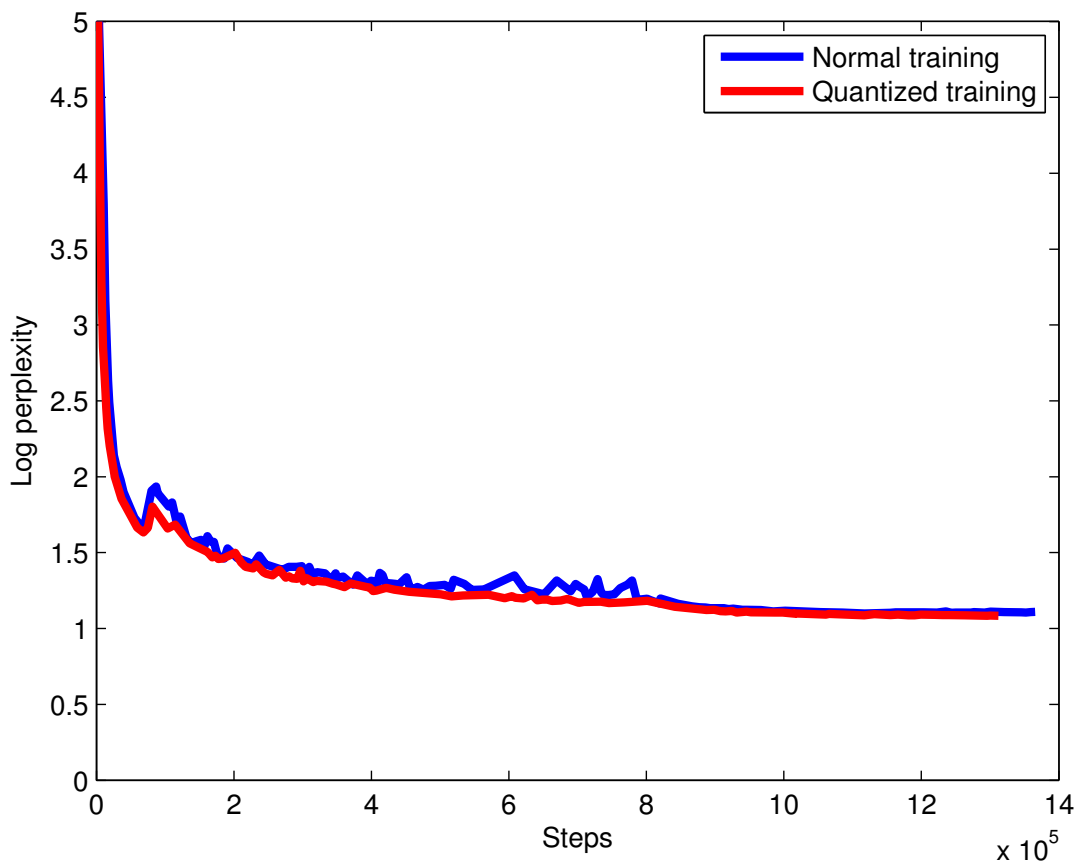}}
	\caption{Log perplexity vs. steps for normal (non-quantized)
          training and quantization-aware training on WMT'14 English
          to French during maximum likelihood training. Notice the
          training losses are similar, with the quantization-aware
          loss being slightly better. Our conjecture for
          quantization-aware training being slightly better is that
          the clipping constraints act as additional regularization which
          improves the model quality.}
\label{quantization_training}
\end{center}
\end{figure}

It is worth emphasizing that during training of the model we use full-precision
floating point numbers.  The only constraints we add to the model
during training are the clipping of the RNN accumulator values into
$[-\delta, \delta]$ and softmax logits into
$[-\gamma, \gamma]$. $\gamma$ is fixed to be at $25.0$, while the
value for $\delta$ is gradually annealed from a generous bound of
$\delta=8.0$ at the beginning of training, to a rather stringent bound
of $\delta=1.0$ towards the end of training. At inference time,
$\delta$ is fixed at $1.0$. Those additional constraints do not degrade
model convergence nor the decoding quality of the model when it has
converged. In Figure \ref{quantization_training}, we compare the loss
vs. steps for an unconstrained model (the blue curve) and a constrained
model (the red curve) on WMT'14 English-to-French. We can see that
the loss for the constrained model is slightly better, possibly due to
regularization roles those constraints play.

Our solution strikes a good balance between efficiency and
accuracy. Since the computationally expensive operations (the matrix
multiplications) are done using 8-bit integer operations, our
quantized inference is quite efficient. Also, since error-sensitive
accumulator values are stored using 16-bit integers, our solution is
very accurate and is robust to quantization errors.

In Table~\ref{table:quantized_infer} we compare the inference speed
and quality when decoding the WMT'14 English-to-French development set (a
concatenation of newstest2012 and newstest2013 test sets for a total
of 6003 sentences) on CPU, GPU and Google's Tensor Processing Unit
(TPU) respectively.\footnote{https://cloudplatform.googleblog.com/2016/05/Google-supercharges-machine-learning-tasks-with-custom-chip.html} The model used here for comparison is trained with
quantization constraints on the ML objective only (i.e., without
reinforcement learning based model refinement). When the model is
decoded on CPU and GPU, it is not quantized and all operations
are done using full-precision floats. When it is decoded on TPU, certain 
operations, such as embedding lookup and attention module, remain on the CPU, 
and all other quantized operations are off-loaded
to the TPU. In all cases, decoding is done on a single machine with
two Intel Haswell CPUs, which consists in total of 88 CPU cores
(hyperthreads). The machine is
equipped with an NVIDIA GPU (Tesla k80) for the experiment with GPU or a single 
Google TPU for the experiment with TPU.

Table~\ref{table:quantized_infer} shows that decoding using reduced
precision arithmetics on the TPU suffers a very minimal loss of 0.0072 on
log perplexity, and no loss on BLEU at all. This result matches
previous work reporting that quantizing convolutional neural
network models can retain most of the model quality.

Table~\ref{table:quantized_infer} also shows that decoding our model on CPU
is actually 2.3 times faster than on GPU.  Firstly, our
dual-CPUs host machine offers a theoretical peak FLOP performance which is more
than two thirds that of the GPU.  Secondly, the beam search
algorithm forces the decoder to incur a non-trivial amount of data
transfer between the host and the GPU at every decoding step. Hence,
our current decoder implementation is not fully utilizing the computation
capacities that a GPU can theoretically offer during inference.

Finally, Table~\ref{table:quantized_infer} shows that decoding on TPUs is
3.4 times faster than decoding on CPUs, demonstrating that quantized arithmetics
is much faster on TPUs than both CPUs or GPUs. 


%

\begin{table}[h!]
\caption{Model inference on CPU, GPU and TPU. The model used here for
	comparison is trained with the ML objective only with
	quantization constraints. Results are obtained by decoding the
	WMT En$\rightarrow$Fr development set on CPU, GPU and TPU
	respectively.}
\label{table:quantized_infer}
\centering
\begin{tabular}{r c c c }
\hline\hline 
 & BLEU & Log Perplexity & Decoding time (s)  \\
	     \hline
	CPU  & 31.20 & 1.4553 & 1322                 \\
	GPU & 31.20 & 1.4553 & 3028 \\
	TPU & 31.21 & 1.4626 & 384 \\
\hline 
\end{tabular}
\end{table}


Unless otherwise noted, we always train and evaluate quantized
models in our experiments. Because there is little difference from a
quality perspective between a model decoded on CPUs and one decoded
on TPUs, we use CPUs to decode for model evaluation during training and 
experimentation and use TPUs to serve production traffic.

\section{Decoder}
\label{decoder}
We use beam search during decoding to find the sequence $Y$
that maximizes a score function $s(Y, X)$ given a trained model. We
introduce two important refinements to the pure max-probability based beam
search algorithm: a coverage penalty~\cite{TuLLLL16} and length
normalization. With length normalization, we aim to account for the
fact that we have to compare hypotheses of different length. Without
some form of length-normalization regular beam search will favor
shorter results over longer ones on average since a negative
log-probability is added at each step, yielding lower (more negative) scores for
longer sentences. We first tried to simply divide
by the length to normalize. We then improved on that original heuristic by dividing by
$length^\alpha$, with $0 < \alpha < 1$ where $\alpha$ is optimized on
a development set ($\alpha\in[0.6-0.7]$ was usually found to be
best). Eventually we designed the empirically-better scoring function
below, which also includes a coverage penalty to favor translations
that fully cover the source sentence according to the attention
module.

More concretely, the scoring function $s(Y, X)$ that we employ to
rank candidate translations is defined as follows:


\begin{equation}
\label{eq-decoder-scorer}
	\begin{split}
	s(Y, X) &= \log(P(Y | X)) / lp(Y) + cp(X;Y) \\
	lp(Y) &= \frac{(5 + |Y|) ^ \alpha}{(5 + 1) ^ \alpha} \\
	cp(X;Y) &= \beta * \sum_{i=1}^{|X|}{\log(\min(\sum_{j=1}^{|Y|}{p_{i, j}}, 1.0))},
	\end{split}
\end{equation}
where $p_{i,j}$ is the attention probability of the $j$-th target word
$y_j$ on the $i$-th source word $x_i$. By construction
(equation \ref{atten}), $\sum_{i=0}^{|X|}{p_{i, j}}$ is equal
to 1.  Parameters $\alpha$ and $\beta$ control the strength of
the length normalization and the coverage penalty. When $\alpha=0$ and
$\beta=0$, our decoder falls back to pure beam search by probability.

During beam search, we typically keep 8-12 hypotheses but we find that
using fewer (4 or 2) has only slight negative effects on BLEU scores. Besides pruning
the number of considered hypotheses, two other forms of pruning are
used. Firstly, at each step, we only consider tokens that have local
scores that are not more than $beamsize$ below the best token for this
step.  Secondly, after a normalized best score has been found
according to equation \ref{eq-decoder-scorer}, we prune all hypotheses
that are more than $beamsize$ below the best normalized score so far.
The latter type of pruning only applies to full hypotheses because it
compares scores in the normalized space, which is only available when
a hypothesis ends. This latter form of pruning also has the effect that
very quickly no more hypotheses will be generated once a sufficiently
good hypothesis has been found, so the search will end quickly. The
pruning speeds up search by $30\%-40\%$ when run on CPUs compared to not
pruning (where we simply stop decoding after a predetermined maximum output
length of twice the source length).  
Typically we use $beamsize=3.0$, unless otherwise noted.

To improve throughput during decoding we can put many sentences (typically
up to 35) of similar length into a batch and decode all of those in parallel to
make use of available hardware optimized for parallel computations. In this
case the beam search only finishes if all hypotheses for all sentences in the
batch are out of beam, which is slightly less efficient theoretically, but in
practice is of negligible additional computational cost.

\begin{table}[h!]
\centering
\begin{tabular}{|r|r|*{6}{r|}}
\hline\hline
\multicolumn{2}{|c|}{}     &   \multicolumn{6}{c|}{$\alpha$} \\ \cline{3-8}
\multicolumn{2}{|c|}{BLEU} &  0.0 &  0.2 &  0.4 &  0.6 &  0.8 &  1.0 \\ \cline{1-8}
             &       0.0   & 30.3 & 30.7 & 30.9 & 31.1 & 31.2 & 31.1 \\ \cline{2-8}
             &       0.2   & 31.4 & 31.4 & 31.4 & 31.3 & 30.8 & 30.3 \\ \cline{2-8}
$\beta$      &       0.4   & 31.4 & 31.4 & 31.4 & 31.1 & 30.5 & 29.6 \\ \cline{2-8}
             &       0.6   & 31.4 & 31.4 & 31.3 & 30.9 & 30.1 & 28.9 \\ \cline{2-8}
             &       0.8   & 31.4 & 31.4 & 31.2 & 30.8 & 29.8 & 28.1 \\ \cline{2-8}
             &       1.0   & 31.4 & 31.3 & 31.2 & 30.6 & 29.4 & 27.2 \\ \cline{2-8}
\hline
\end{tabular}
\caption{WMT'14 En$\rightarrow$Fr BLEU score with respect to different values of $\alpha$ and
  $\beta$. The model in this experiment trained using ML without RL
	refinement. A single WMT En$\rightarrow$Fr model achieves a
	BLEU score of 30.3 on the development set when the beam search scoring
	function is purely based on the sequence probability (i.e.,
	both $\alpha$ and $\beta$ are $0$). Slightly larger $\alpha$
	and $\beta$ values improve BLEU score by up to $+1.1$
	($\alpha=0.2,\beta=0.2$), with a wide range of $\alpha$ and $\beta$ 
        values giving results very close to the best BLEU scores.}
\label{table:decoder}
\end{table}

Table \ref{table:decoder} shows the impact of $\alpha$ and $\beta$ on
the BLEU score when decoding the WMT'14 English-to-French development set.
The model used here for experiments is trained using the ML objective
only (without RL refinement). As can be seen from the results, having
some length normalization and coverage penalty improves BLEU score
considerably (from 30.3 to 31.4).

We find that length normalization ($\alpha$) and coverage penalty
($\beta$) are less effective for models with RL
refinement. Table \ref{table:decoder_after_rl} summarizes our
results. This is understandable, as during RL refinement, the models
already learn to pay attention to the full source sentence to not
under-translate or over-translate, which would result in a penalty on the
BLEU (or GLEU) scores.

\begin{table}[h!]
\centering
\begin{tabular}{|r|r|*{6}{r|}}
\hline\hline
\multicolumn{2}{|c|}{}     &   \multicolumn{6}{c|}{$\alpha$} \\ \cline{3-8}
\multicolumn{2}{|c|}{BLEU} &  0.0 &  0.2 &  0.4 &  0.6 &  0.8 &  1.0 \\ \cline{1-8}
	&0.0 & 0.320 & 0.321 & 0.322 & 0.322 & 0.322 & 0.322 \\ \cline{2-8}
	&0.2 & 0.322 & 0.322 & 0.322 & 0.322 & 0.321 & 0.321 \\ \cline{2-8}
$\beta$ &0.4 & 0.322 & 0.322 & 0.322 & 0.321 & 0.321 & 0.316 \\ \cline{2-8}
	&0.6 & 0.322 & 0.322 & 0.321 & 0.321 & 0.319 & 0.309 \\ \cline{2-8}
	&0.8 & 0.322 & 0.322 & 0.321 & 0.321 & 0.316 & 0.302 \\ \cline{2-8}
	&1.0 & 0.322 & 0.321 & 0.321 & 0.320 & 0.313 & 0.295 \\ \cline{2-8}
\hline
\end{tabular}
\caption{WMT En$\rightarrow$Fr BLEU score with respect to different
  values of $\alpha$ and $\beta$. The model used here is trained using ML, then
  refined with RL. Compared to the results in Table \ref{table:decoder},
  coverage penalty and length normalization appear to be less
  effective for models after RL-based model refinements. Results are 
  obtained on the development set.}
\label{table:decoder_after_rl}
\end{table}

We found that the optimal $\alpha$ and $\beta$ vary slightly for 
  different models. Based on tuning results using internal Google
  datasets, we use $\alpha=0.2$ and $\beta=0.2$ in our experiments, unless
  noted otherwise.

\section{Experiments and Results}
\label{Experimental Setup}
In this section, we present our experimental results on
two publicly available corpora used extensively as
benchmarks for Neural Machine Translation systems:
WMT'14 English-to-French (WMT En$\rightarrow$Fr) and
English-to-German (WMT En$\rightarrow$De). On these two datasets, we
benchmark GNMT models with word-based, character-based, and wordpiece-based
vocabularies. We also present the improved accuracy of our models after
fine-tuning with RL and model ensembling. Our main objective
with these datasets is to show the contributions of various components
in our implementation, in particular the wordpiece model, RL
model refinement, and model ensembling.

In addition to testing on publicly available corpora, we also test GNMT on
Google's translation production corpora, which are two to three decimal orders of magnitudes bigger than the WMT corpora for a given language pair. We
compare the accuracy of our model against human accuracy and the 
best Phrase-Based Machine Translation (PBMT) production system for Google Translate.

In all experiments, our models consist of 8 encoder layers and 8 decoder layers.
(Since the bottom encoder layer is actually bi-directional, in total there are
9 logically distinct LSTM passes in the encoder.)
The attention network is a simple feedforward network with one hidden layer with 1024 nodes.
All of the models use 1024 LSTM nodes per encoder and decoder layers.


\subsection{Datasets}
We evaluate our model on the WMT En$\rightarrow$Fr dataset, the WMT
En$\rightarrow$De dataset, as well as many Google-internal
production datasets. On WMT En$\rightarrow$Fr, the training set
contains 36M sentence pairs.  On WMT En$\rightarrow$De, the training
set contains 5M sentence pairs. In both cases, we use newstest2014 as the test
sets to compare against previous
work~\cite{luong2015addressing,jean2015using,DBLP:journals/corr/ZhouCWLX16}.
The combination of newstest2012 and newstest2013 is used as the development set.

In addition to WMT, we also evaluate
our model on some Google-internal datasets representing a wider
spectrum of languages with distinct linguistic properties:
English $\leftrightarrow$ French, English $\leftrightarrow$ Spanish and
English $\leftrightarrow$ Chinese.

\subsection{Evaluation Metrics}
We evaluate our models using the standard BLEU score metric. To be
comparable to previous work~\cite{sutskever2014sequence,luong2015addressing,DBLP:journals/corr/ZhouCWLX16}, we report
tokenized BLEU score as computed by the \texttt{multi-bleu.pl} script,
downloaded from the public implementation of Moses (on Github), which is
also used in~\cite{luong2015addressing}.

As is well-known, BLEU score does not fully capture the quality of a
translation. For that reason we also carry out side-by-side (SxS)
evaluations where we have human raters evaluate and compare the
quality of two translations presented side by side for a given source
sentence. Side-by-side scores range from 0 to 6, with a score of 0
meaning \textsl{``completely nonsense translation''}, and a score of 6
meaning \textsl{``perfect translation: the meaning of the translation
is completely consistent with the source, and the grammar is
correct''}. A translation is given a score of 4 if \textsl{``the
sentence retains most of the meaning of the source sentence, but may
have some grammar mistakes''}, and a translation is given a score of 2
if \textsl{``the sentence preserves some of the meaning of the source
sentence but misses significant parts''}. These scores are generated
by human raters who are fluent in both languages and hence often
capture translation quality better than BLEU scores.



\subsection{Training Procedure}
The models are trained by a system we implemented using
TensorFlow\cite{tensorflow16}.
The training setup follows the classic
data parallelism paradigm. There are 12 replicas running
concurrently on separate machines. Every replica updates the shared
parameters asynchronously.

We initialize all trainable parameters uniformly between [-0.04, 0.04]. As
is common wisdom in training RNN models, we apply gradient clipping
(similar to \cite{sutskever2014sequence}): all gradients are uniformly
scaled down such that the norm of the modified gradients is no larger
than a fixed constant, which is $5.0$ in our case. If the norm of the
original gradients is already smaller than or equal to the given
threshold, then gradients are not changed.

For the first stage of maximum likelihood training (that is, to
optimize for objective function \ref{eq:objml}), we use a
combination of Adam \cite{DBLP:journals/corr/KingmaB14} and simple SGD
learning algorithms provided by the TensorFlow runtime system.  We run Adam for
the first 60k steps, after which we switch to simple SGD. Each step in
training is a mini-batch of 128 examples. 

\begin{figure}[h!]
\begin{center}
\centerline{\includegraphics[width=\textwidth]{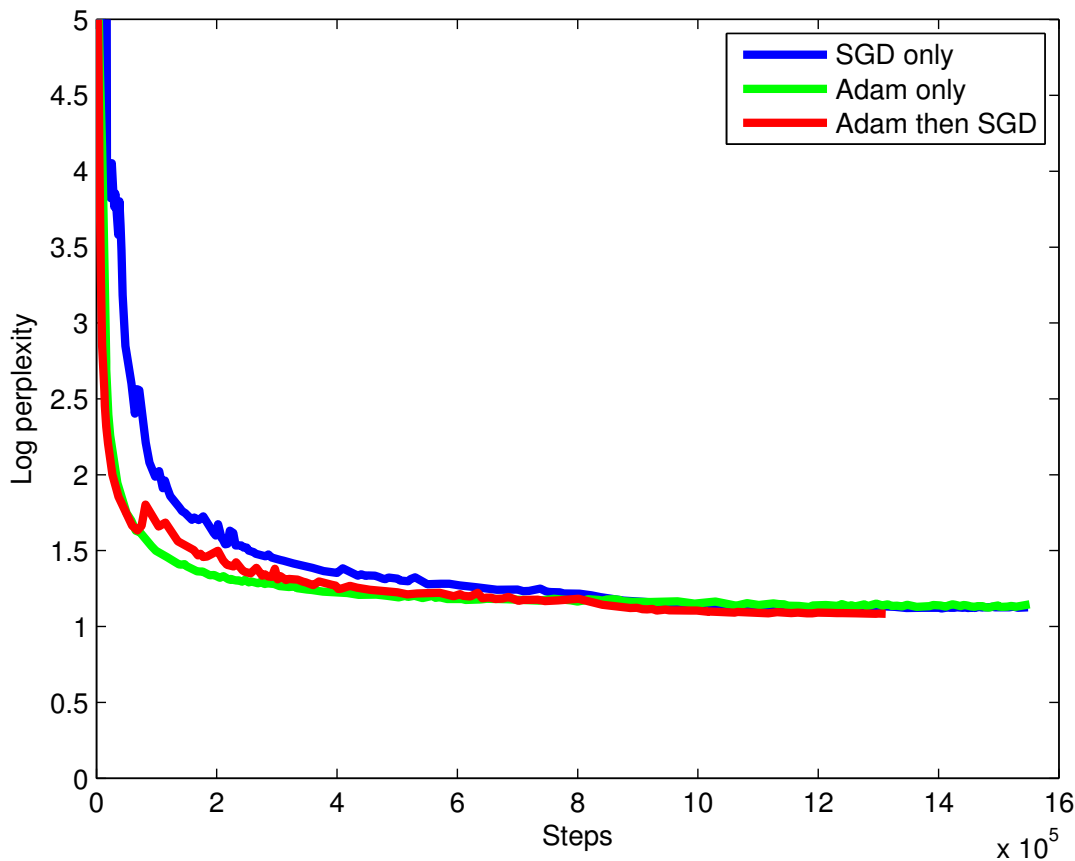}}
	\caption{Log perplexity vs. steps for Adam, SGD and
          Adam-then-SGD on WMT En$\rightarrow$Fr during maximum
          likelihood training.  Adam converges much faster than SGD at
          the beginning.  Towards the end, however, Adam-then-SGD is
          gradually better. Notice the bump in the red curve
          (Adam-then-SGD) at around 60k steps where we switch from
          Adam to SGD.  We suspect that this bump occurs due to
          different optimization trajectories of Adam vs. SGD. When we
          switch from Adam to SGD, the model first suffers a little,
          but is able to quickly recover afterwards.}
\label{sgd_adam}
\end{center}
\end{figure}

We find that Adam accelerates training at the beginning, but Adam
alone converges to a worse point than a combination of Adam first, followed by 
SGD (Figure~\ref{sgd_adam}). For the Adam part, we use a learning
rate of $0.0002$, and for the SGD part, we use a learning rate of
$0.5$. We find that it is important to also anneal the learning rate
after a certain number of total steps. For the WMT En$\rightarrow$Fr dataset, 
we begin
to anneal the learning rate after 1.2M steps, after which we halve the
learning rate every 200k steps for an additional 800k steps. On WMT
En$\rightarrow$Fr, it takes around 6 days to train a basic model using 96 NVIDIA K80 GPUs.

%


Once a model is fully converged using the ML objective, we switch to RL
based model refinement, i.e., we further optimize the objective
function as in equation \ref{eq:mixed}. We refine a model until the BLEU
score does not change much on the development set.  For this model
refinement phase, we simply run the SGD optimization algorithm. The number
of steps needed to refine a model varies from dataset to dataset. For
WMT En$\rightarrow$Fr, it takes around 3 days to complete 400k steps. 

To prevent overfitting, we apply dropout during training with a scheme similar
to \cite{lstm_dropout}. For the WMT En$\rightarrow$Fr and En$\rightarrow$De datasets, we set the
dropout probability to be $0.2$ and $0.3$ respectively. Due to various technical reasons, dropout is
only applied during the ML training phase, not during the RL refinement phase.

The exact hyper-parameters vary from dataset to dataset and from model
to model. For the WMT En$\rightarrow$De dataset, since it is significantly
smaller than the WMT En$\rightarrow$Fr dataset, we use a higher dropout
probability, and also train smaller models for fewer steps overall. 
On the production data sets, we typically do not use dropout, and
we train the models for more steps.

\subsection{Evaluation after Maximum Likelihood Training}

The models in our experiments are word-based, character-based, mixed
word-character-based or several wordpiece models with varying
vocabulary sizes.

For the word model, we selected the most frequent 212K source words as
the source vocabulary and the most popular 80k target words as the
target vocabulary. Words not in the source vocabulary or the target
vocabulary (unknown words) are converted into special 
\texttt{<first\_char>\_UNK\_<last\_char>}
symbols. Note, in this case, there is more than one UNK (e.g., our
production word models have roughly 5000 different UNKs in this case). We then
use the attention mechanism to copy a corresponding word from the
source to replace these unknown words during decoding~\cite{jean2015using}.

The mixed word-character model is similar to the word model, except the
out-of-vocabulary (OOV) words are converted into sequences of
characters with special delimiters around them as described in section
\ref{Mixed Word/Character Model} in more detail. In
our experiments, the vocabulary size for the mixed word-character
model is 32K. For the pure character model, we simply split all words
into constituent characters, resulting typically in a few hundred basic
characters (including special symbols appearing in the data). For the
wordpiece models, we train 3 different models with vocabulary sizes of
8K, 16K, and 32K.



Table \ref{table:wmt_en_fr} summarizes our results on the WMT
En$\rightarrow$Fr dataset. In this table, we also compare against other
strong baselines without model ensembling.  As can be seen from the
table, ``WPM-32K'', a wordpiece model with a shared source and target
vocabulary of 32K wordpieces, performs well on this dataset and achieves the
best quality as well as the fastest inference speed.

The pure character model (char input, char output) works surprisingly
well on this task, not much worse than the best wordpiece models in BLEU 
score. However, these models are rather slow to train and slow to use as the 
sequences are much longer.

Our best model, WPM-32K, achieves a BLEU score of 38.95. Note that this
BLEU score represents the averaged score of 8 models we trained. The maximum
BLEU score of the 8 models is higher at 39.37.  We point out
that our models are completely self-contained, as opposed to previous models
reported in~\cite{DBLP:journals/corr/ZhouCWLX16}, which depend on
some external alignment models to achieve their best results. Also note that
all our test set numbers were achieved by picking an optimal model on the
development set which was then used to decode the test set.

Note that the timing numbers for this section are obtained on CPUs, not TPUs.
We use here the same CPU machine as described above, and run the decoder with
a batchsize of 16 sentences in parallel and a maximum of 4 concurrent
hypotheses at any time per sentence. The time per sentence is the total
decoding time divided by the number of respective sentences in the test set.




\begin{table}[h!]
\caption{Single model results on WMT En$\rightarrow$Fr (newstest2014)}
\label{table:wmt_en_fr}
\centering
\begin{tabular}{r c c c }
\hline\hline 
       Model     & BLEU        & CPU decoding time    \\
                 &             & per sentence (s) \\ \hline
       Word      & 37.90       & 0.2226           \\
       Character & 38.01       & 1.0530           \\
       WPM-8K    & 38.27       & 0.1919           \\
       WPM-16K   & 37.60       & 0.1874           \\
       WPM-32K   & 38.95       & 0.2118           \\

       Mixed Word/Character& 38.39 & 0.2774    \\\hline
       PBMT~\cite{durrani2014edinburgh} & 37.0  & \\
       LSTM (6 layers)~\cite{luong2015addressing} & 31.5   & \\
       LSTM (6 layers + PosUnk)~\cite{luong2015addressing} & 33.1   & \\
       Deep-Att~\cite{DBLP:journals/corr/ZhouCWLX16} & 37.7 &  \\
       Deep-Att + PosUnk~\cite{DBLP:journals/corr/ZhouCWLX16} & 39.2  & \\
\hline 
\end{tabular}
\end{table}


Similarly, the results of WMT En$\rightarrow$De are presented in
Table~\ref{table:wmt_en_de}. Again, we find that wordpiece models
achieves the best BLEU scores.

\begin{table}[h!]
\caption{Single model results on WMT En$\rightarrow$De (newstest2014)} 
\label{table:wmt_en_de}
\centering
\begin{tabular}{r c c }
\hline\hline 
       Model     & BLEU        & CPU decoding time    \\
                 &             & per sentence (s) \\ \hline
       Word                  & 23.12   & 0.2972           \\
       Character (512 nodes) & 22.62   & 0.8011           \\
       WPM-8K                & 23.50   & 0.2079           \\
       WPM-16K               & 24.36   & 0.1931           \\
       WPM-32K               & 24.61   & 0.1882           \\
       Mixed Word/Character  & 24.17   & 0.3268           \\ 
       \hline
       PBMT~\cite{buck2014n}                           & 20.7     & \\
       RNNSearch~\cite{jean2015using} & 16.5            \\
       RNNSearch-LV~\cite{jean2015using} & 16.9          \\
       RNNSearch-LV~\cite{jean2015using} & 16.9          \\
       Deep-Att~\cite{DBLP:journals/corr/ZhouCWLX16} & 20.6    & \\
\hline 
\end{tabular}
\end{table}


WMT En$\rightarrow$De
is considered a more difficult task than WMT En$\rightarrow$Fr as
it has much less training data, and German, as a more morphologically
rich language, needs a huge vocabulary for word models.  Thus it is
more advantageous to use wordpiece or mixed word/character models,
which provide a gain of more than 2 BLEU points on top of the
word model and about 4 BLEU points on top of previously reported results
in~\cite{buck2014n,DBLP:journals/corr/ZhouCWLX16}.
Our best model, WPM-32K, achieves a BLEU score of 24.61, which is averaged over 8 runs.
Consistently, on the production corpora, wordpiece models tend
to be better than other models both in terms of speed and accuracy.


\subsection{Evaluation of RL-refined Models}
The models trained in the previous section are optimized for
log-likelihood of the next step prediction
which may not correlate well with translation quality, as discussed in
section~\ref{sec:RL}. We use RL training to fine-tune sentence BLEU scores
after normal maximum-likelihood training.

The results of RL fine-tuning on the best En$\rightarrow$Fr and
En$\rightarrow$De models are presented in
Table~\ref{table:rl_wmt_en_fr}, which show that fine-tuning the
models with RL can improve BLEU scores. On WMT En$\rightarrow$Fr,
model refinement improves BLEU score by close to 1 point. On En$\rightarrow$De,
RL-refinement slightly hurts the test performance even though we observe about 0.4 BLEU points
improvement on the development set. The results presented in
Table~\ref{table:rl_wmt_en_fr} are the average of 8 independent models.
We also note that there is an overlap between the wins from the RL refinement and the decoder 
fine-tuning (i.e., the introduction of length normalization and coverage penalty).
On a less fine-tuned decoder (e.g., if the decoder does beam search by
log-probability only), the win from RL would have been bigger (as is evident
from comparing results in Table~\ref{table:decoder} and
Table~\ref{table:decoder_after_rl}).

\begin{table}[h!]
\caption{Single model test BLEU scores, averaged over 8 runs, on WMT En$\rightarrow$Fr and
  En$\rightarrow$De}
\label{table:rl_wmt_en_fr}
\centering
\begin{tabular}{r c c c }
\hline\hline 
       Dataset & Trained with log-likelihood   & Refined with RL  \\\hline
       En$\rightarrow$Fr  & 38.95                  & 39.92           \\ \hline
       En$\rightarrow$De  & 24.67                  & 24.60           \\
\hline 
\end{tabular}
\end{table}

\subsection{Model Ensemble and Human Evaluation}
We ensemble 8 RL-refined models to obtain a state-of-the-art
result of 41.16 BLEU points on the WMT En$\rightarrow$Fr dataset. Our results are
reported in Table~\ref{table:ensemble-wmt_en_fr}.
\begin{table}[h!]
\caption{Model ensemble results on WMT En$\rightarrow$Fr (newstest2014)}
\label{table:ensemble-wmt_en_fr}
\centering
\begin{tabular}{r c c }
\hline\hline 
       Model                                                             & BLEU        \\
       \hline
       WPM-32K (8 models)                                                & 40.35 \\
       RL-refined WPM-32K (8 models)                                     & 41.16 \\ \hline
       LSTM (6 layers)~\cite{luong2015addressing}                        & 35.6 \\
       LSTM (6 layers + PosUnk)~\cite{luong2015addressing}               & 37.5 \\
       Deep-Att + PosUnk (8 models)~\cite{DBLP:journals/corr/ZhouCWLX16} & 40.4 \\
\hline 
\end{tabular}
\end{table}

We ensemble 8 RL-refined models to obtain a state-of-the-art
result of 26.30 BLEU points on the WMT En$\rightarrow$De dataset. Our results are
reported in Table~\ref{table:ensemble-wmt_en_de}.
\begin{table}[h!]
\caption{Model ensemble results on WMT En$\rightarrow$De (newstest2014). See Table~\ref{table:wmt_en_de} for a comparison against non-ensemble models.}
\label{table:ensemble-wmt_en_de}
\centering
\begin{tabular}{r c c }
\hline\hline 
       Model                                                             & BLEU        \\
       \hline
       WPM-32K (8 models)                                                & 26.20 \\
       RL-refined WPM-32K (8 models)                                     & 26.30 \\
\hline 
\end{tabular}
\end{table}

Finally, to better understand the quality of our models and the effect
of RL refinement, we carried out a four-way side-by-side human
evaluation to compare our NMT translations against the reference translations
and the best phrase-based statistical machine translations.
During the side-by-side comparison,
humans are asked to rate four translations given a source sentence.
The four translations are:
1) the best phrase-based translations as downloaded
from \textsl{http://matrix.statmt.org/systems/show/2065},
2) an ensemble of 8 ML-trained models,
3) an ensemble of 8 ML-trained and then RL-refined models, and
4) reference human translations as taken directly from newstest2014,
Our results are presented in Table~\ref{table:wmt_en_fr_sxs}.

\begin{table}[h!]
\caption{Human side-by-side evaluation scores of WMT En$\rightarrow$Fr models.}
\centering
\begin{tabular}{r c c}
\hline\hline 
Model & BLEU & Side-by-side \\
      &      & averaged score \\ \hline
	PBMT~\cite{durrani2014edinburgh} & 37.0 & 3.87 \\ %
	NMT before RL & 40.35 & 4.46 \\
	NMT after RL & 41.16 & 4.44 \\ \hline %
	Human &  & 4.82 \\ %
\hline 
\end{tabular}
\label{table:wmt_en_fr_sxs}
\end{table}

The results show that even though RL refinement can achieve better
BLEU scores, it barely improves the human impression of the translation
quality.  This could be due to a combination of factors including: 1)
the relatively small sample size for the experiment (only 500
examples for side-by-side), 2) the improvement in BLEU score by RL
is relatively small after model ensembling (0.81), which may be at a
scale that human side-by-side evaluations are insensitive to, and 3) the
possible 
mismatch between BLEU as a metric and real translation quality as perceived by
human raters. Table~\ref{table:example_translations} contains some example
translations from PBMT, "NMT before RL" and "Human", along with the
side-by-side scores that human raters assigned to each translation
(some of which we disagree with, see the table caption).




\subsection{Results on Production Data}

We have carried out extensive experiments on many Google-internal production
data sets.
As the experiments above cast doubt on whether RL improves the real translation
quality or simply the BLEU metric, RL-based model
refinement is not used during these experiments.
Given the larger volume of training data available in the Google corpora,
dropout is also not needed in these experiments.

\begin{table}[h!]
\caption{Mean of side-by-side scores on production data} 
\centering
\begin{tabular}{l c c c c }
\hline\hline 
	& PBMT & GNMT & Human & Relative\\ [0.5ex]
	&      &      &       & Improvement\\ [0.5ex]
\hline 
English $\rightarrow$ Spanish     & 4.885 & 5.428 & 5.504 & 87\%\\ %
English $\rightarrow$ French	  & 4.932 & 5.295 & 5.496 & 64\%\\ %
English $\rightarrow$ Chinese     & 4.035 & 4.594 & 4.987 & 58\%\\ %
Spanish $\rightarrow$ English     & 4.872 & 5.187 & 5.372 & 63\%\\ %
French  $\rightarrow$ English	  & 5.046 & 5.343 & 5.404 & 83\%\\ %
Chinese $\rightarrow$ English     & 3.694 & 4.263 & 4.636 & 60\%\\ %
\hline 
\end{tabular}
\label{table:prod}
\end{table}

\begin{figure}[h!]
\centering
\includegraphics[width=6in]{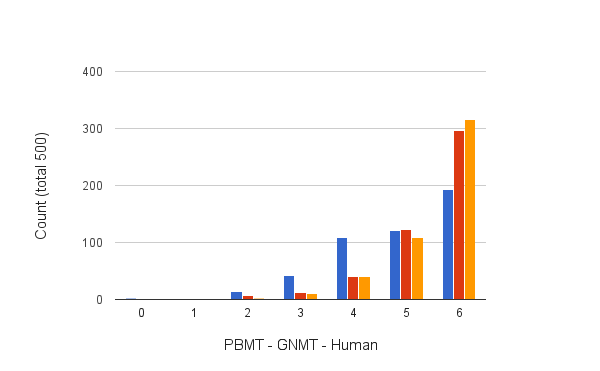}
\caption{Histogram of side-by-side scores on 500 sampled sentences from Wikipedia and news websites for a typical language pair, here English $\rightarrow$ Spanish (PBMT blue, GNMT red, Human orange). It can be seen that there is a wide distribution in scores, even for the human translation when rated by other humans, which shows how ambiguous the task is. It is clear that GNMT is much more accurate than PBMT.}
\label{fig:histogram}
\end{figure}

In this section we describe our experiments with human perception of the
translation quality. We asked human raters to rate translations in
a three-way side-by-side comparison.  The three sides are from: 1) translations 
from the production phrase-based statistical translation system used by Google, 
2) translations from our GNMT system, and 3) translations by humans fluent in 
both languages.  Reported here in Table \ref{table:prod} are averaged rated 
scores for English
$\leftrightarrow$ French, English $\leftrightarrow$ Spanish and
English $\leftrightarrow$
Chinese. All the GNMT models are wordpiece models, without model
ensembling, and use a shared source and target vocabulary with 32K wordpieces. 
On each pair of languages, the evaluation data consist of 500
randomly sampled sentences from Wikipedia and news websites, and the
corresponding human translations to the target language. The
results show that our model reduces translation errors by more than 60\%
compared to the PBMT model on these major pairs of languages. A typical
distribution of side-by-side scores is shown in Figure \ref{fig:histogram}.

As expected, on this metric the GNMT system improves also compared to the PBMT
system. In some cases human and GNMT translations are
nearly indistinguishable on the relatively simplistic and isolated sentences
sampled from Wikipedia and news articles for this experiment. Note that we have
observed that human raters, even though fluent in both languages, do not
necessarily fully understand each randomly sampled sentence sufficiently and 
hence cannot necessarily generate the best possible translation or rate a 
given translation accurately. Also note that, although the scale for the
scores goes from
0 (complete nonsense) to 6 (perfect translation) the human translations
get an imperfect score of only around 5 in Table \ref{table:prod}, which shows
possible ambiguities in the translations and also possibly non-calibrated raters
and translators with a varying level of proficiency.

Testing our GNMT system on particularly difficult translation cases and longer
inputs than just single sentences is the subject of future work.





\section{Conclusion}
In this paper, we describe in detail the implementation of Google's Neural 
Machine Translation (GNMT) system, including all the techniques that are critical 
to its accuracy, speed, and robustness.
On the public WMT'14 translation benchmark, our system's translation quality
approaches or surpasses all currently published results.
More importantly, we also show that our approach carries over to much larger
production data sets, which have several orders of magnitude more data, to 
deliver high quality translations.

Our key findings are:
1) that wordpiece modeling effectively handles open vocabularies
and the challenge of morphologically rich languages for translation quality 
and inference speed,
2) that a combination of model and data parallelism can be used to efficiently
train state-of-the-art sequence-to-sequence NMT models in roughly a week,
3) that model quantization drastically accelerates translation inference,
allowing the use of these large models in a deployed production environment, and
4) that many additional details like length-normalization, coverage penalties,
and similar are essential to making NMT systems work well on real data.

Using human-rated side-by-side comparison as a metric, we show that our
GNMT system approaches the accuracy achieved by average bilingual human 
translators on some of our test sets.
In particular, compared to the previous phrase-based production system,
this GNMT system delivers roughly a 60\% reduction in translation errors
on several popular language pairs.

\section*{Acknowledgements}
We would like to thank the entire Google Brain Team and Google Translate Team
for their foundational contributions to this project.


\bibliography{bnmt}
\bibliographystyle{acm}

\newpage
\appendix
\newcolumntype{L}[1]{>{\raggedright\let\newline\\\arraybackslash\hspace{0pt}}m{#1}}

\begin{table}
	\caption{Some example translations from PBMT~\cite{durrani2014edinburgh}, our GNMT system (the "NMT before RL", Table~\ref{table:wmt_en_fr_sxs}), and Human. Source and target sentences (human translations) are from the public benchmark WMT En$\rightarrow$Fr (newstest2014) data set. The right-hand column shows the human ratings on a scale of 0 (complete nonsense) to 6 (perfect translation). We disagree with some of the human ratings, e.g., the translation \emph{``Elle a été repéré trois jours plus tard par un promeneur de chien piégé dans la carrière''} contains grammatical mistakes and changes semantics, and is still scored 6. We present it to illustrate the potential problems of the scoring process.}

\label{table:example_translations}
\centering
	\begin{tabular}{ c  L{\dimexpr.8\textwidth}  c }
\noalign{\vskip 2mm}
\hline\hline 
		Source & "The reason Boeing are doing this is to cram more seats in to make their plane more competitive with our products," said Kevin Keniston, head of passenger comfort at Europe's Airbus. & \\
		\hline
		PBMT & "La raison pour laquelle Boeing sont en train de faire, c'est de concentrer davantage de sièges pour prendre leur avion plus compétitive avec nos produits", a déclaré Kevin M. Keniston, chef du confort des passagers de l'Airbus de l'Europe. &  3.0 \\
		\hline
		GNMT & "La raison pour laquelle Boeing fait cela est de créer plus de sièges pour rendre son avion plus compétitif avec nos produits", a déclaré Kevin Keniston, chef du confort des passagers chez Airbus. & 6.0 \\
		\hline
		Human & "Boeing fait ça pour pouvoir caser plus de sièges et rendre ses avions plus compétitifs par rapports à nos produits", a déclaré Kevin Keniston, directeur de Confort Passager chez l'avionneur européen Airbus. & 6.0 \\\hline\hline
		Source & When asked about this, an official of the American administration replied: "The United States is not conducting electronic surveillance aimed at offices of the World Bank and IMF in Washington." & \\\hline
		PBMT & Interrogé à ce sujet, un responsable de l'administration américaine a répondu : "Les Etats-Unis n'est pas effectuer une surveillance électronique destiné aux bureaux de la Banque mondiale et du FMI à Washington". & 3.0 \\\hline
		GNMT & Interrogé à ce sujet, un fonctionnaire de l'administration américaine a répondu: "Les États-Unis n'effectuent pas de surveillance électronique à l'intention des bureaux de la Banque mondiale et du FMI à Washington". &  6.0 \\\hline
		Human & Interrogé sur le sujet, un responsable de l'administration américaine a répondu: "les Etats-Unis ne mènent pas de surveillance électronique visant les sièges de la Banque mondiale et du FMI à Washington". & 6.0 \\\hline\hline
		Source&Martin told CNN that he asked Daley whether his then-boss knew about the potential shuffle. & \\\hline
		PBMT&Martin a déclaré à CNN qu'il a demandé Daley si son patron de l'époque connaissaient le potentiel remaniement ministériel.& 2.0\\\hline
		GNMT&Martin a dit à CNN qu'il avait demandé à Daley si son patron d'alors était au courant du remaniement potentiel.& 6.0 \\\hline
		Human&Martin a dit sur CNN qu'il avait demandé à Daley si son patron d'alors était au courant du remaniement éventuel.& 5.0 \\\hline\hline
		Source&She was spotted three days later by a dog walker trapped in the quarry& \\\hline
		PBMT&Elle a été repéré trois jours plus tard par un promeneur de chien piégé dans la carrière& 6.0 \\\hline
		GNMT&Elle a été repérée trois jours plus tard par un traîneau à chiens piégé dans la carrière.& 2.0 \\\hline
		Human&Elle a été repérée trois jours plus tard par une personne qui promenait son chien coincée dans la carrière& 5.0 \\\hline\hline

		Source&Analysts believe the country is unlikely to slide back into full-blown conflict, but recent events have unnerved foreign investors and locals. & \\\hline
		PBMT&Les analystes estiment que le pays a peu de chances de retomber dans un conflit total, mais les événements récents ont inquiété les investisseurs étrangers et locaux.&5.0\\\hline
		GNMT&Selon les analystes, il est peu probable que le pays retombe dans un conflit généralisé, mais les événements récents ont attiré des investisseurs étrangers et des habitants locaux.&2.0\\\hline
		Human&Les analystes pensent que le pays ne devrait pas retomber dans un conflit ouvert, mais les récents évènements ont ébranlé les investisseurs étrangers et la population locale. &5.0\\\hline\hline

\hline 
\end{tabular}
\end{table}

\end{document}